\documentclass[runningheads]{llncs}
\usepackage[T1]{fontenc}
\usepackage{graphicx,verbatim}
\usepackage{amsfonts}

\usepackage{xcolor}
\usepackage[colorlinks=true, citecolor=blue, linkcolor=black, urlcolor=magenta]{hyperref}

\usepackage{amsmath}
\usepackage{xcolor}

\definecolor{darkblue}{RGB}{0, 0, 255} 
\definecolor{darkred}{RGB}{128, 0, 0} 
\definecolor{darkgreen}{RGB}{0, 100, 0} 
\definecolor{darkpurple}{RGB}{128, 0, 128} 
\definecolor{darkorange}{RGB}{204, 85, 0} 
\definecolor{darkgray}{RGB}{255, 0, 0} 

\usepackage{tikz,xcolor,hyperref}

\definecolor{lime}{HTML}{A6CE39}
\DeclareRobustCommand{\orcidicon}{
	\begin{tikzpicture}
	\draw[lime, fill=lime] (0,0) 
	circle [radius=0.16] 
	node[white] {{\fontfamily{qag}\selectfont \tiny ID}};
	\draw[white, fill=white] (-0.0625,0.095) 
	circle [radius=0.007];
	\end{tikzpicture}
	\hspace{-2mm}
}

\foreach \x in {A, ..., Z}{\expandafter\xdef\csname orcid\x\endcsname{\noexpand\href{https://orcid.org/\csname orcidauthor\x\endcsname}
			{\noexpand\orcidicon}}
}

\usepackage{tikz}
\usepackage{xparse}   

\DeclareRobustCommand{\authorpic}[2][5mm]{%
  \tikz[baseline={([yshift=-.25ex]current bounding box.center)}]{%
    \clip (0,0) circle (#1);
    \pgfmathsetlengthmacro{\picside}{sqrt(2)*#1}%
    \node at (0,0) {\includegraphics[width=\picside,height=\picside,keepaspectratio]{#2}};
    \draw[line width=0.4pt, color=white] (0,0) circle (#1);
  }%
}

\NewDocumentCommand{\AuthorWithPic}{O{5.5mm} O{0.20em} m m}{%
  \texorpdfstring{\authorpic[#1]{#4}\kern #2}{}%
  #3%
}

\usepackage{xcolor}
\definecolor{linkpinkix}{HTML}{EA335A} 

\definecolor{linkpink}{HTML}{EA335A}

\newcommand{\shadedlink}[2]{%
  \tikz[baseline=(n.base)]\node[
    fill=linkpink,
    fill opacity=0.5,
    text opacity=1,
    rounded corners=.3ex,
    inner xsep=.35em,
    inner ysep=.15em
  ] (n) {\href{#1}{\textcolor{blue!70!black}{#2}}};%
}

\begin{document}
\title{
\textcolor{darkblue}{N}\textcolor{darkred}{E}\textcolor{darkgreen}{U}\textcolor{darkpurple}{R}\textcolor{darkorange}{A}\textcolor{darkgray}{L}:\
Attentio\textcolor{darkblue}{n}-Guided \ 
Pruning \ 
for \ 
Unifi\textcolor{darkred}{e}d \ 
M\textcolor{darkgreen}{u}ltimodal \ 
\textcolor{darkpurple}{R}esource-Constr\textcolor{darkorange}{a}ined
Clinical \ 
Eva\textcolor{darkgray}{l}uation
}
%

\author{%
  \AuthorWithPic[6mm][0.18em]{Devvrat Joshi}{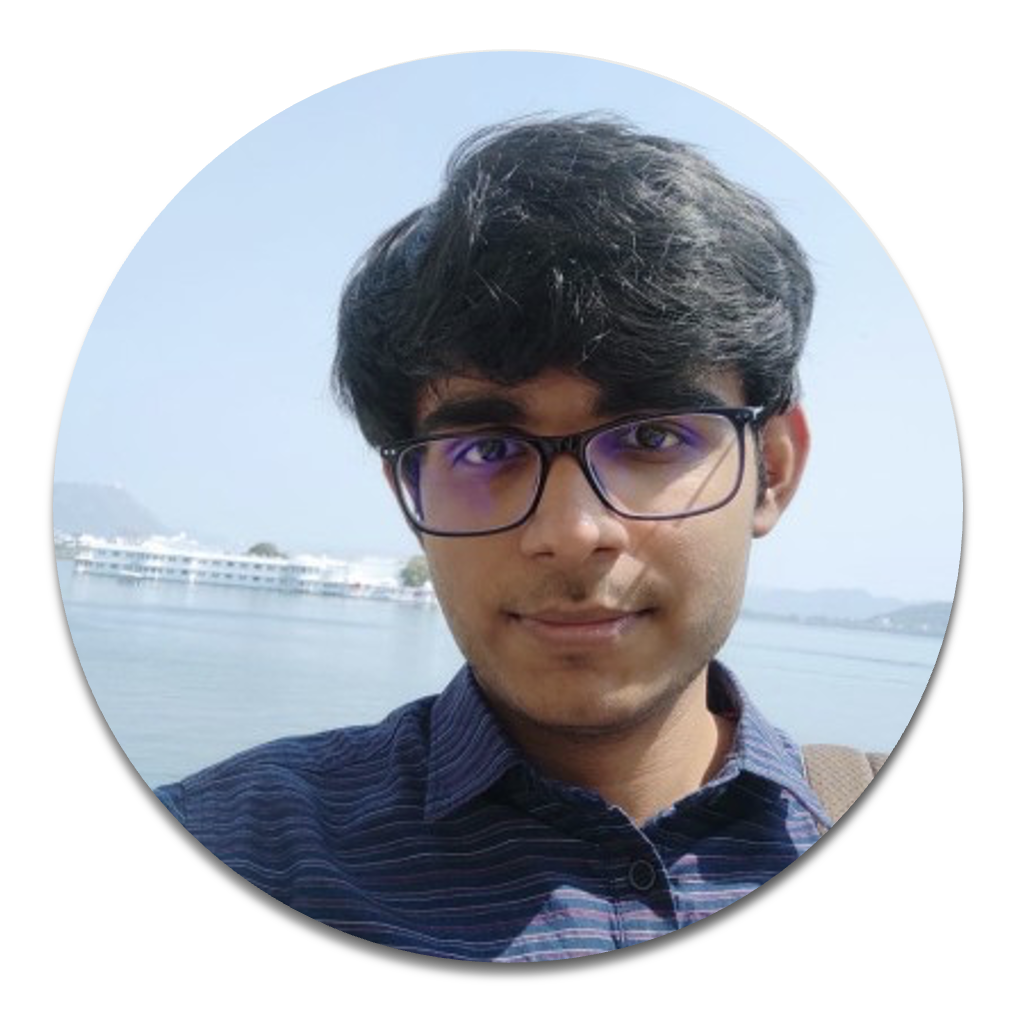} \and
  \AuthorWithPic[6mm][0.18em]{Islem Rekik\orcidA{}}{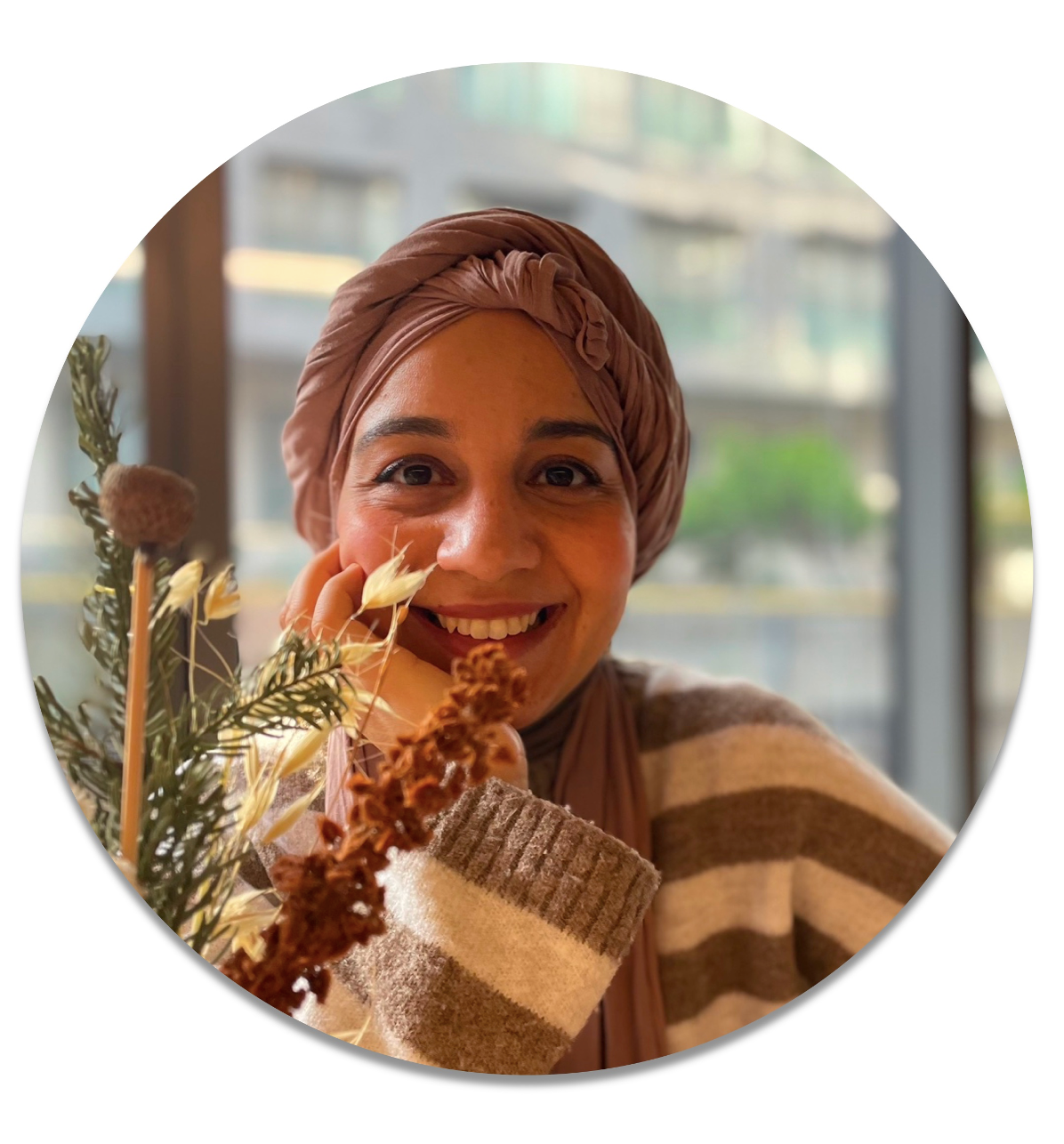}\thanks{Corresponding author: \email{i.rekik@imperial.ac.uk}, \url{http://basira-lab.com}, GitHub: \url{https://github.com/basiralab/NEURAL}}%
}

\institute{BASIRA Lab, Imperial-X (I-X) and Department of Computing, Imperial College London, London, United Kingdom} 



\maketitle              
\begin{abstract}

The rapid growth of multimodal medical imaging data presents significant storage and transmission challenges, particularly in resource-constrained clinical settings. We propose NEURAL, a novel framework that addresses this by using semantics-guided data compression. Our approach repurposes cross-attention scores between the image and its radiological report from a fine-tuned generative vision-language model to structurally prune chest X-rays, preserving only diagnostically critical regions. This process transforms the image into a highly compressed, graph representation. This unified graph-based representation fuses the pruned visual graph with a knowledge graph derived from the clinical report, creating a universal data structure that simplifies downstream modeling. Validated on the MIMIC-CXR and CheXpert Plus dataset for pneumonia detection, NEURAL achieves a 93.4-97.7\% reduction in image data size while maintaining a high diagnostic performance of 0.88-0.95 AUC, outperforming other baseline models that use uncompressed data. By creating a persistent, task-agnostic data asset, NEURAL resolves the trade-off between data size and clinical utility, enabling efficient workflows and teleradiology without sacrificing performance. Our NEURAL code is available at \href{https://github.com/basiralab/NEURAL}{https://github.com/basiralab/NEURAL}.\footnote{This paper has been selected for an \textbf{Oral Presentation} at the CLIP MICCAI 2025 workshop. \shadedlink{https://youtu.be/6GZ_Gpk1KZM}{[NEURAL YouTube Video]}.}


\end{abstract}

\vspace{-0.5cm}
\begin{figure}
    \centering
    \includegraphics[width=1\linewidth]{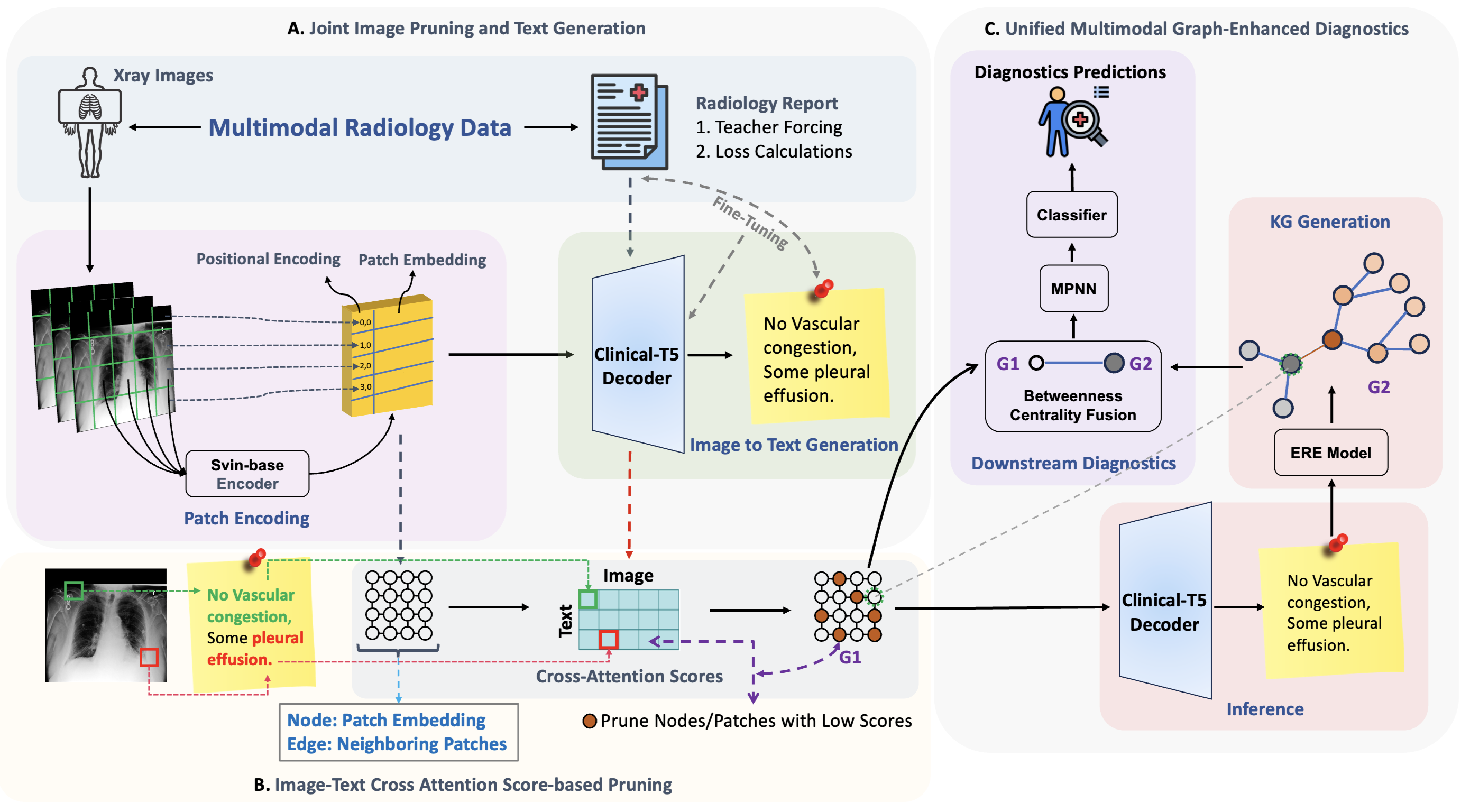}
    \caption{End-to-end NEURAL pipeline for report-guided image pruning and graph-based clinical diagnostics.}
    \label{fig:main-figure}
\end{figure}

\section{Introduction}

The volume of medical imaging data is expanding rapidly, with over two billion chest X-rays (CXRs) performed annually worldwide \cite{akhter2023ai}. Each exam typically includes both an image and a radiology report, resulting in massive multimodal datasets that can reach terabytes in size. This data is vital for training AI models and supporting clinical workflows, but its scale creates serious challenges in real-world deployment. Many hospitals, especially in low-resource settings, face storage limitations, slow networks, and limited computing power \cite{magudia2021trials}. These challenges delay timely image interpretation, making it difficult to integrate AI tools into clinical practice, and limit the reach of remote diagnostic services. Without effectively addressing these barriers, the full benefits of medical imaging AI may remain unrealized in real-world clinical settings \cite{proof_of_work}.

Current efforts to manage this data complexity largely follow two distinct paradigms, each with fundamental limitations. The first is model-centric, focusing on accelerating computation. This ranges from established network pruning methods like CheXPrune \cite{kaur2022chexprune} to more recent, sophisticated techniques that use language guidance to dynamically discard irrelevant tokens during inference \cite{sun2025lvpruningeffectivesimplelanguageguided,cao-etal-2023-pumer}. While these approaches can significantly reduce the computational footprint of a model, their objective is transient model acceleration, not persistent data reduction. They still expect access to the original, full-resolution image for every task, thereby failing to alleviate the core issues of data storage and transmission. The second paradigm, conventional image compression, uses methods like JPEG to reduce file size but remains agnostic to clinical content, risking the degradation or loss of diagnostically critical details \cite{Fischer2023}. This presents a stark trade-off: one can either optimize the model while leaving the data logistics problem unsolved, or compress the data at the risk of compromising its clinical integrity.

To resolve this trade-off, we propose a novel framework centered on semantics-guided data compression. Our approach leverages the rich clinical narrative of the radiology report as a semantic blueprint to guide a targeted, structural compression of the associated image. Unlike methods that first downsize the image and risk losing information, our framework operates on the \textit{full-resolution} image data, ensuring fine-grained visual details are considered during the pruning process. This identifies and preserves diagnostically critical regions while systematically discarding redundant information. The final output is not merely a compressed image with its radiological report, but a structured, multimodal representation in the form of a graph, which can be serialized into a lightweight format like a pickle file for efficient and lossless transmission.

This graph-based representation acts as a universal data structure that is inherently extensible. It is designed to seamlessly incorporate additional data types in the future, such as temporal clinical data or MRI scans, and to effectively model the complex interactions between them. Crucially, as we show in section \ref{methodology} (Parts B and C), this unification into a graph format eliminates the need for complex, task-specific models that handle heterogeneous inputs. Instead, a single, much simpler graph-based model can be applied for diverse downstream tasks, achieving comparable performance to other methods while operating on the highly compressed data.

Our pipeline operates on image text pairs from the MIMIC-CXR \cite{Johnson_Pollard_Berkowitz_Greenbaum_Lungren_Deng_Mark_Horng_2019} and CheXpert Plus \cite{chambon2024chexpertplusaugmentinglarge} dataset. Each image is first represented as a graph of visual patches (Refer Section \ref{methodology}-Part A). During fine-tuning, a ClinicalT5 \cite{lu-etal-2022-clinicalt5} decoder is trained to generate the corresponding ground truth report from these visual inputs. The core innovation of our approach is to repurpose the cross-attention scores, calculated between full resolution image patches and clinical text tokens, as explicit signals for structurally pruning the image-graphs. We then evaluate the fidelity and clinical utility of the resulting compressed graph through two downstream tasks: (1) radiology report generation and (2) pneumonia classification. As a result, this work makes four key contributions:

1. We introduce the \textbf{first semantics-guided framework for radiological image compression} that uses cross modal attention from a generative vision language model to explicitly guide structural pruning of medical images, enabling highly targeted compression. 

2. Our compression framework is \textbf{task-agnostic}, enabling the compressed data to be stored and utilized across downstream applications. 

3. We develop a \textbf{rigorous dual-validation strategy}, using both report generation and disease classification to evaluate the fidelity and clinical relevance of the compressed representations.

4. Our framework is \textbf{designed for future extensibility}, leveraging a powerful betweenness centrality fusion to create a unified graph. This strategy is highly efficient, connecting modalities via a single semantically-meaningful link to avoid quadratic complexity , which in turn allows for the natural incorporation of new data types and simplifies downstream models.

\vspace{-0.3cm}
\section{Related Work}

While NEURAL shares a core mechanism with contemporary language-guided pruning methods like LVPruning \cite{sun2025lvpruningeffectivesimplelanguageguided} and PuMer \cite{cao-etal-2023-pumer}, their foundational objectives and task-dependencies diverge significantly. The primary goal of these other approaches is transient model acceleration, where pruning is intrinsically linked to a specific, concurrent downstream task. For instance, in an instruction-following model like IVTP \cite{10.1007/978-3-031-72643-9_13} and ATP-LLaVA \cite{ye2024atpllavaadaptivetokenpruning}, the visual tokens are pruned based on the immediate query, making the pruning a consequence of that single task. NEURAL’s objective, in contrast, is persistent data compression. It aims to create a smaller, permanent, and task-agnostic data asset that addresses the more fundamental challenges of data storage and transmission, particularly within resource-constrained clinical environments.

This distinction in purpose leads to a crucial divergence in methodology and generalizability. For other methods, the pruning is an ephemeral part of an inference pipeline, re-calculated for each new task or prompt. As a result, a different clinical application requires pruning of the full image specific to that task. NEURAL’s methodology, however, is designed to be performed once; it uses the holistic clinical narrative of a ground-truth report to create a single, static, compressed graph. This resulting data asset is inherently versatile. Because the pruning is guided by the comprehensive report rather than a narrow downstream task, the compressed graph is a general-purpose representation that can be used for any number of subsequent clinical applications, be it pneumonia classification, report generation, or other diagnostic queries, without modification.

\vspace{-0.4cm}
\section{Methodology}\label{methodology}
\vspace{-0.2cm}

Our framework proposes a semantics-guided, 3-stage approach to compress medical images and unify multimodal data into a single graph representation for downstream applications. The process begins by dividing a full-resolution chest X-ray into non-overlapping patches. Subsequently, a generative vision-language model is fine-tuned to create a radiology report, generating cross-attention scores that link text to the image patches. These scores are then repurposed to structurally prune the image, distilling it into a sparse graph containing only the most salient visual regions. Finally, this pruned visual graph is fused with a knowledge graph derived from the clinical report, creating the unified multimodal graph for efficient downstream diagnostics.

\textbf{A) Joint Image Pruning and Report Generation.}\label{part_a} 
A central challenge in multimodal medical AI is bridging the semantic gap between the dense, low-level pixel data of a radiograph and the sparse, high-level concepts expressed in a clinical report. Our methodology addresses this by training an encoder-decoder module for a dual purpose: not only to generate coherent radiology reports but also, through this process, to produce a fine-grained alignment map identifying the most clinically salient image regions. This transforms the standard task of report generation into a tool for extracting the most important semantically aware visual regions in the image.

Our approach begins by processing a chest radiograph (CXR), \( I \in \mathbb{R}^{H \times W \times 3} \). Following the Vision Transformer (ViT) paradigm \cite{liu2021swintransformerhierarchicalvision}, the image is divided into a sequence of \( N \) non-overlapping patches. These patches are linearly embedded, along with 2-dimensional positional embeddings for retaining global context, and fed into a Swin encoder \cite{liu2021swintransformerhierarchicalvision}. This produces a sequence of patch-level feature representations, \( V_{\text{img}} = \{v_1, v_2, \dots, v_N\} \), where each \( v_i \in \mathbb{R}^{D_{\text{vis}}} \) captures visual information from a specific image region. To make these features compatible with our language model, we project them into the text embedding space using a dedicated linear projection layer, resulting in the final visual embeddings \( E_{\text{vis}} \in \mathbb{R}^{N \times D_{\text{text}}} \).

The core innovation lies in how we leverage these visual embeddings. We fine-tune a pre-trained clinical language model decoder (\texttt{Clinical-T5-Base}) \cite{lu-etal-2022-clinicalt5} on report generation, conditioning it directly on the visual embeddings \( E_{\text{vis}} \). During fine-tuning, we employ a teacher-forcing strategy: given an image \( I \) and its ground-truth report \( R = \{t_1, t_2, \dots, t_M\} \), the decoder learns to predict each token \( t_j \) based on the full set of visual embeddings \( E_{\text{vis}} \) and the preceding ground-truth tokens \( \{t_1, \dots, t_{j-1}\} \). This training encourages the model to establish direct, meaningful correlations between textual concepts and the specific image patches supporting them. A critical byproduct is the cross-attention mechanism, which produces attention scores at each decoding step quantifying the importance of each image patch for generating each token. Unlike conventional uses of attention for feature fusion, we repurpose this dynamic, context-aware attention map as a precise, data-driven signal to guide structural pruning of the visual graph, as detailed in the following section.

\textbf{B) Image-Text Cross-Attention Score-based Pruning. }In medical imaging, treating all parts of an image with equal importance creates a large volume of redundant data. This conventional method is inefficient and can obscure critical diagnostic clues within the noise. Our work introduces a new approach in radiological image analysis by leveraging the powerful semantic connection between an image and its medical report to intelligently filter this information. This process distills the dense image graph into a sparse, meaningful subgraph. By doing so, we create a focused map containing only the most clinically relevant visual evidence, leading to more efficient and accurate diagnostic models.

To facilitate this, the pruning mechanism repurposes the cross-attention scores generated during the report generation phase described previously. For each token \( t_j \) in the ground-truth report \( R \), the decoder produces an attention weight vector, \(\boldsymbol{\alpha}_j = \{\alpha_{j,1}, \alpha_{j,2}, \dots, \alpha_{j,N}\}\), where the scalar \(\alpha_{j,i}\) quantifies the importance of the \(i\)-th image patch, \(v_i\), to generating that specific token. To determine the overall relevance of each patch to the entire clinical narrative, we aggregate these scores across all tokens. The cumulative importance score \( S_i \) for each patch \( v_i \) is computed as: $ S_i = \sum_{j=1}^{M} \alpha_{j,i} $. This aggregated score \( S_i \) serves as a robust proxy for the clinical salience of the corresponding image region.

We define a threshold, $\tau$, and construct a new, pruned set of vertices, $V'_{img}$, by retaining only those nodes whose cumulative attention scores exceed this threshold: $
V'_{img} = \{v_i \in V_{img} \mid S_i > \tau \}
$. The threshold \(\tau\) can be determined empirically or set dynamically to retain a top-\(k\) percentage of the most salient patches, providing precise control over the desired level of compression. The resulting pruned graph, denoted as \(G_1\) in Figure 1, is a semantically compressed representation of the original image. It is no longer a generic grid of patches but a customized data structure shaped by the clinical text to retain only the most critical visual information, thereby reducing data size and focusing subsequent analyses on diagnostically significant regions.

\textbf{C) Unified Multimodal Graph-Enhanced Diagnostics.} The final stage of our framework aims for a holistic diagnostic capability, \textit{moving beyond superficial feature fusion toward true structural integration of modalities.} We unify our pruned visual graph, $G_1$, with a structured representation of the clinical text. Simply using the raw report would overlook the rich interdependencies between medical concepts and image patches, and would require a multimodal model for handling text and graph simultaneously. We therefore transform the clinical narrative into a textual Knowledge Graph (KG), $G_2$. Using a BiomedVLP-CXR-BERT model used in \cite{delbrouck-etal-2024-radgraph}, we extract medical findings and entities as nodes and represent their semantic relationships as edges. This structured representation allows the model to reason explicitly about how clinical concepts relate, rather than processing the report as a flat sequence of tokens. 

We construct a knowledge graph $G_2$ from the report, and fuse it with the visual graph $G_1$, and then pass the combined structure through a MPNN for reasoning. The fusion is performed by connecting the nodes with the highest betweenness centrality from each graph, creating a semantically meaningful link between the two modalities. While other approaches, such as connecting all nodes based on cross-attention scores with edge weights, are possible, we choose to add only one edge to avoid the quadratic increase in complexity. This design supports efficient structural fusion, promoting interpretability and enabling a more context-aware multimodal diagnostic process.

To reason over this heterogeneous graph, we experimented with both homogeneous and heterogeneous graph neural network architectures. While heterogeneous GNNs preserve modality-specific semantics more explicitly, we found that their added complexity and training overhead did not yield performance gains significant enough to justify their use in this context. Instead, we adopt a standard Message Passing Neural Network (MPNN). Its iterative message-passing mechanism enables each node, whether an image patch or a clinical concept, to refine its representation based on both its neighborhood and its cross-modal links. This modeling choice strikes a practical balance between computational efficiency and representational richness.

At inference time, we consider two options: using the decoder to generate the text or directly using the textual report from the dataset. The choice depends on the specific requirements of the clinical setting, whether the hospital prefers to manage image-report pairs or rely solely on a compressed image graph for inference. 

\textbf{Why Graphs?} Operating over a graph data structure offers significant advantages for both data storage and downstream tasks. First, it enables future extensibility by providing a unified framework that can naturally incorporate additional modalities, such as clinical temporal data and MRI, facilitating the modeling of interactions between heterogeneous data sources. Second, it standardizes inputs for downstream models, allowing them to work with a consistent graph format rather than a complex mixture of images, text, and temporal signals, thereby simplifying the input pipeline. Finally, the use of a cross-attention mechanism allows for efficient learning by projecting multimodal representations onto a shared low-dimensional manifold, enabling even simple message-passing neural networks (MPNNs) to effectively learn decision boundaries with relatively few parameters.

\vspace{-0.3cm}

\section{Experiments and Discussion}\label{experiments}
We conduct a comprehensive set of experiments to validate our proposed framework. Our evaluation is designed to answer three critical questions: (1) How effectively our method performs after data compression compared to established baselines? (2) Does the compressed graph representation retain sufficient clinical information for high-fidelity downstream tasks? (3) How does the quality of the textual guidance impact the trade-off between compression and diagnostic accuracy? We demonstrate that our approach achieves an unprecedented level of compression while maintaining state-of-the-art diagnostic performance, and we analyze the key factors influencing its behavior through extensive ablation studies.

\textbf{Datasets.} Our framework was rigorously evaluated using two distinct multimodal radiology datasets: MIMIC-CXR \cite{Johnson_Pollard_Berkowitz_Greenbaum_Lungren_Deng_Mark_Horng_2019} and CheXpert Plus \cite{chambon2024chexpertplusaugmentinglarge}. To ensure the integrity of our results and prevent data leakage, we enforced a strict patient-level separation, guaranteeing that only one imaging study per patient was included, thus preventing any patient's data from appearing in both training and testing splits.

The primary benchmark for classification was based on the MIMIC-CXR dataset, comprising 377,000 chest X-ray images. From this corpus, we identified 40,894 images with definitive pneumonia labels (24,338 negative and 16,556 positive), yielding a positive sample ratio of 40.5\%. To better reflect the class imbalance typically seen in clinical practice, we constructed a more challenging dataset by sampling 10,000 of these labeled images to create a distribution with a pneumonia prevalence of only 15\%. This allowed for a rigorous evaluation of model performance in detecting a sparsely represented target class. For external validation and assessment of generalization, we additionally utilized the CheXpert dataset. From its full set of 223,462 radiographs, we selected the 1,296 images that were explicitly labeled for pneumonia, providing a focused test set drawn from a distinct patient population. Finally, we divide the datasets into training (70\%), validation (15\%) and testing (15\%) sets.

\textbf{Baselines.} We evaluated our framework for generating radiological reports using two other models. The RGRG \cite{Tanida_2023} method detects anatomical regions in chest X-rays and generates region-specific sentences grounded on predicted bounding boxes, enhancing explainability and interactivity. CvT2DistilGPT2 \cite{NICOLSON2023102633} improves report generation by warm-starting its encoder with a Convolutional vision Transformer (CvT-21) pre-trained on ImageNet-21K and its decoder with DistilGPT2, resulting in more accurate and radiologist-like reports. 

For the pneumonia detection task, we chose models that utilize both text as well as images for a fair comparison with NEURAL: CheXMed \cite{REN2024119854} is a multimodal algorithm for pneumonia detection that fuses features extracted from X-ray images via CNN and clinical notes processed through Named Entity Recognition into a combined representation for classification. RMT (Robust Multimodal Transformer) \cite{Li_Nan_Qi_Cai_Zhao_Li_Liu_Wang_Wu_Miao_etal._2024} assesses pediatric pneumonia severity by integrating X-rays and medical records using a Transformer architecture with multi-task learning and mask attention to handle missing data, achieving superior performance in multimodal settings.

\textbf{Compression and Diagnostic Performance.} Our framework strikes an effective balance between extreme data compression and high clinical accuracy, outperforming traditional approaches.
As shown in Table~\ref{tab:main_results}, our method achieves a 97.7\% reduction in data size by pruning the image graph down to just 2.3\% on MIMIC-CXR dataset and 93.4\% reduction in CheXPert dataset of its original nodes. Despite this significant compression, the AUC remains high at 0.947 and 0.875 for MIMIC-CXR and CheXPert datasets, surpassing the performance of other multimodal approaches that combine text and image inputs. While we initially use the radiology report generator to learn cross attention scores for pruning, it can also be leveraged to generate reports, eliminating the need to store original radiology texts. However, generating reports from pruned nodes does lead to a drop in performance, as reflected in lower AUC scores.

\vspace{-0.5cm}

\begin{table}[h!]
\centering
\caption{BLEU-2 Scores for Radiology Report Generation.}
\label{tab:bleu_scores_bordered}
\begin{tabular}{|l|c|c|c|c|}
\hline
\textbf{Model} & \textbf{MIMIC-CXR}& \textbf{CheXpert} & \textbf{Model Params} & \textbf{Image Resolution}\\
\hline
RGRG \cite{Tanida_2023}  & 0.21 & 0.15 & 220M & 512$\times$512\\
\hline
CvT2DistilGPT2 \cite{NICOLSON2023102633} & 0.09 & 0.08 & 102M & 384$\times$384\\
\hline
\textbf{NEURAL (Ours)} & \textbf{0.23} & \textbf{0.18} & 308M & Full\\
\hline
\textbf{NEURAL (Pruned)} & 0.19 & 0.16 & 308M & Pruned\\
\hline
\end{tabular}
\label{tab:bleu}
\end{table}

\textbf{Report Generation Quality.} Although report generation is primarily used to extract cross attention scores for pruning, we also assess the quality of the generated reports to ensure the language model remains coherent. As shown in Table~\ref{tab:bleu}, our fine-tuned Clinical-T5 model achieves strong performance on the BLEU-2 metric, comparable to other models of similar size~\cite{Tanida_2023,NICOLSON2023102633}. This suggests that our pruning strategy is guided by clinically meaningful and coherent text generation. In contrast to prior works that reduces image resolution, our patch-based method operates on full-resolution images, enabling the encoder to retain fine-grained visual details. We also evaluate Clinical-T5 on the reduced patch set, using a compression ratio similar to that in Table~\ref{tab:main_results}. While the BLEU-2 scores drop relative to the full data, the model still performs reasonably well, indicating some loss of coherence due to the pruned patches that have low cross-attention scores.

\begin{table}[h!]
\centering
\caption{Pneumonia Detection Performance (AUC) Across Datasets. CI references to Max-Compression on images, GT refers to the use of Generated Text. Results are defined as AUC vs \% Compression}
\label{tab:auc_comparison_bordered}
\begin{tabular}{|l|c|c|}
\hline
\textbf{Model} & \textbf{MIMIC-CXR} & \textbf{CheXpert} \\
\hline
CheXMed \cite{REN2024119854} & 0.939, 0\% & 0.816, 0\% \\
\hline
RMT \cite{Li_Nan_Qi_Cai_Zhao_Li_Liu_Wang_Wu_Miao_etal._2024} & 0.915, 0\% & 0.869, 0\% \\
\hline
\textbf{NEURAL (No Pruning)} &\textbf{0.963}, 0\% & \textbf{0.902}, 0\% \\
\hline
\textbf{NEURAL (CI)} & 0.947, 97.7\% & 0.875, 93.4\% \\
\hline
\textbf{NEURAL (CI + GT)} & 0.891, 97.7\% & 0.838, 93.4\% \\
\hline
\end{tabular}
\label{tab:main_results}
\end{table}

\begin{figure}[h]
    \centering
    \begin{minipage}[b]{0.49\textwidth}
        \centering
        \includegraphics[width=\textwidth]{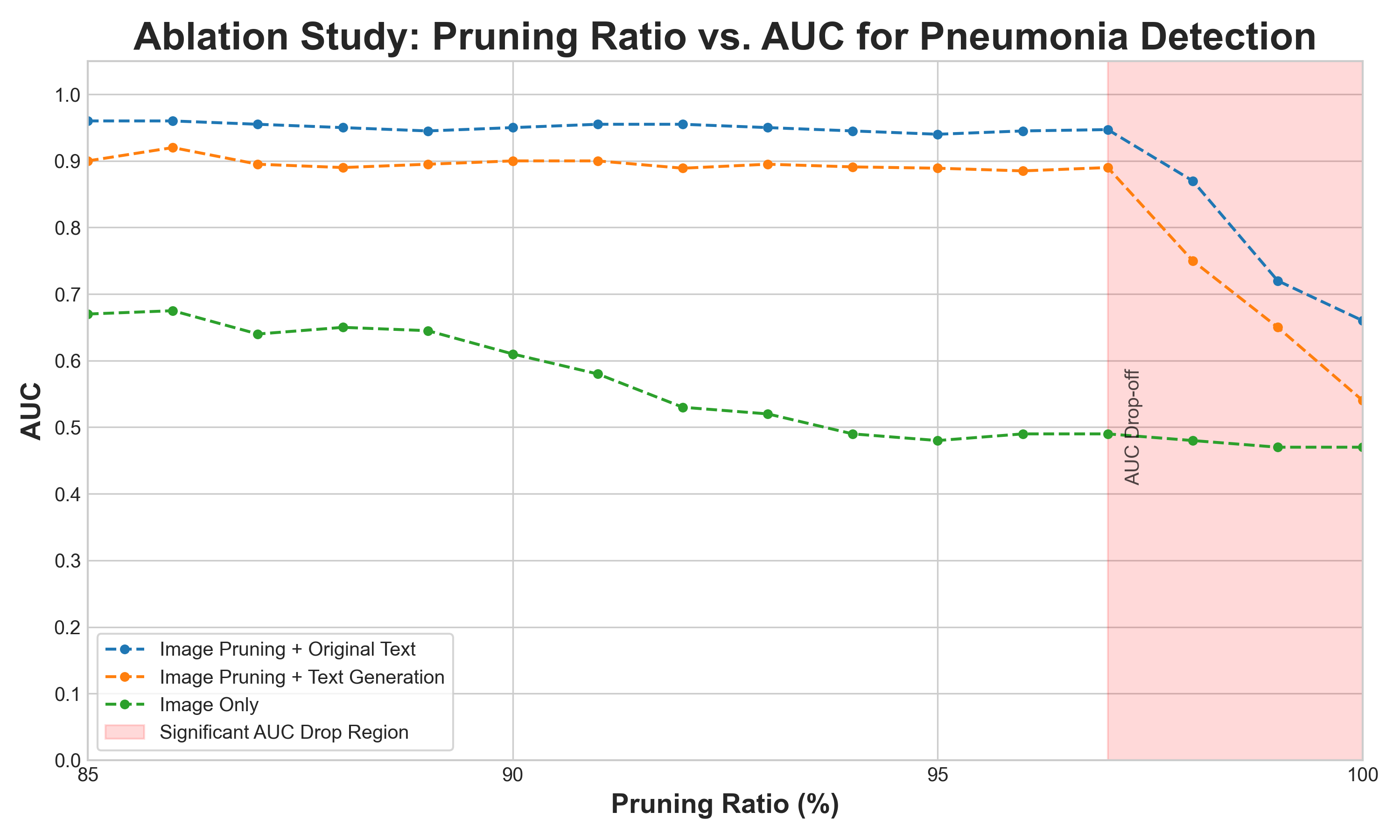}
    \end{minipage}
    \hfill
    \begin{minipage}[b]{0.50\textwidth}
        \centering
        \includegraphics[width=\textwidth]{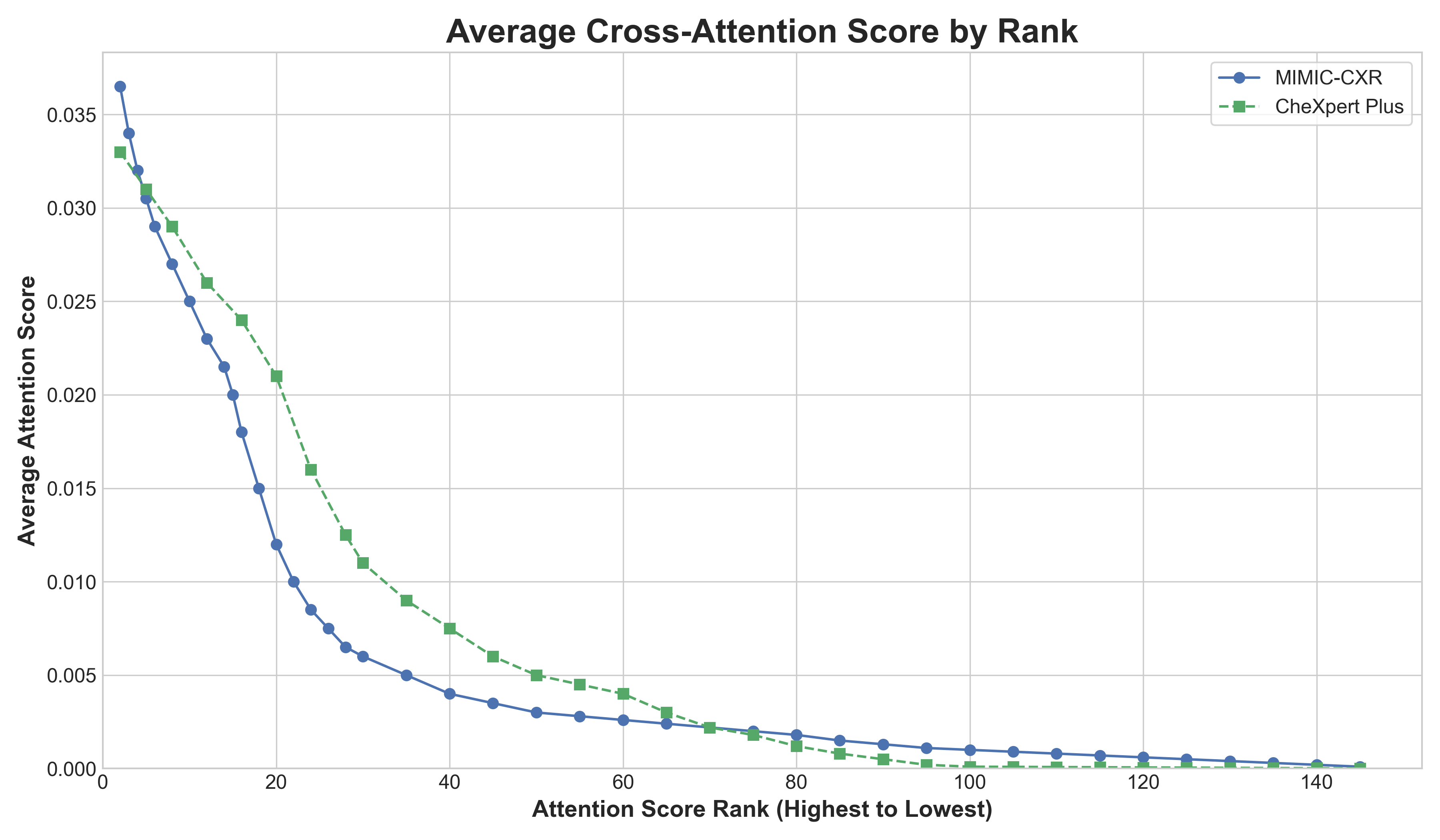}
    \end{minipage}
    \caption{Figure on the left shows the ablation study of varying the pruning ratio on three seperate tasks for MIMIC-CXR dataset. Figure on the right shows the average cross-attention scores vs rank of the patch inside the image for the CheXpert and MIMIC-CXR datasets}
    \label{fig:fig2}
\end{figure}

\textbf{Ablation Study. }We analyze how varying the pruning ratio affects diagnostic accuracy. As shown in Figure~\ref{fig:fig2} (Left), retaining 2.3\% of image patches already yields a strong AUC of 0.95. Increasing the retention to 10\% results in only a slight gain (+0.01 AUC), indicating diminishing returns. This justifies our design choice: 97.7\% compression delivers substantial efficiency with minimal impact on performance. Moreover, we demonstrated that removing text information entirely for prediction, combined with pruning over 90\% of the data, leads to a substantial drop in performance. This highlights the critical role of text tokens in pneumonia detection. Since the other two models do not support unimodal input like NEURAL, we were unable to evaluate their performance using only image data.

To better understand why selecting the top 2.3\% of patches is effective, we analyzed the cross-attention scores of the image patches. For each image, we ranked the patches in descending order based on their cross-attention scores. Then, for each rank, we computed the average score across all images and visualized the results in Figure \ref{fig:fig2} (Right). The plot reveals that the top 20 ranked patches exhibit significantly higher attention scores, which then drop sharply to near zero. These 20 patches correspond to the top 2.3\% patches across the MIMIC-CXR dataset. Similarly, in the CheXpert Plus dataset, a larger proportion of patches exhibit high attention scores, resulting in less aggressive compression compared to the MIMIC-CXR dataset.

\vspace{-0.3cm}
\section{Clinical Implications and Conclusion}
\vspace{-0.2cm}
Our method enables a 93–97\% reduction in image size with minimal loss in task performance, offering significant advantages for clinical imaging workflows. This compression greatly reduces storage demands in Picture Archiving and Communication Systems (PACS) and supports efficient teleradiology in low-bandwidth or resource-constrained environments. The pruned image graphs can be stored in lightweight formats (e.g., pickle) and reused for a variety of downstream tasks. Since the pruning is driven by a task-agnostic report generation model, the approach remains generalizable and adaptable to multiple clinical applications. Unlike prior work that reduces image resolution, our patch-based method preserves full image detail, enabling better retention of fine-grained visual features.

In addition, our fine-tuned Clinical-T5 model can generate coherent and clinically meaningful radiology reports directly from the compressed inputs. This reduces the reliance on storing large, paired image–report datasets and simplifies data management in clinical research and deployment settings. Although some drop in report quality is observed when using heavily pruned inputs, the model still performs reasonably well, demonstrating the effectiveness of our pruning strategy. Overall, this enables scalable report generation with minimal overhead. In future work, we aim to extend this framework to other data types, such as temporal or volumetric imaging, and to design models that can operate directly on compressed graph representations. This would move NEURAL toward a more efficient, generalizable, and end-to-end solution for multimodal clinical data analysis.
\vspace{-0.3cm}
%
%
%
%
%
%

\bibliographystyle{splncs04}
\bibliography{bibliography.bib}

\begin{thebibliography}{10}
\providecommand{\url}[1]{\texttt{#1}}
\providecommand{\urlprefix}{URL }
\providecommand{\doi}[1]{https://doi.org/#1}

\bibitem{akhter2023ai}
Akhter, Y., Singh, R., Vatsa, M.: Ai-based radiodiagnosis using chest x-rays: A review. Frontiers in Big Data  \textbf{6},  1120989 (2023). \doi{10.3389/fdata.2023.1120989}

\bibitem{cao-etal-2023-pumer}
Cao, Q., Paranjape, B., Hajishirzi, H.: {P}u{M}er: Pruning and merging tokens for efficient vision language models. In: Rogers, A., Boyd-Graber, J., Okazaki, N. (eds.) Proceedings of the 61st Annual Meeting of the Association for Computational Linguistics (Volume 1: Long Papers). pp. 12890--12903. Association for Computational Linguistics, Toronto, Canada (Jul 2023). \doi{10.18653/v1/2023.acl-long.721}

\bibitem{chambon2024chexpertplusaugmentinglarge}
Chambon, P., Delbrouck, J.B.: Chexpert plus: Augmenting a large chest x-ray dataset with text radiology reports, patient demographics and additional image formats (2024), \url{https://arxiv.org/abs/2405.19538}

\bibitem{delbrouck-etal-2024-radgraph}
Delbrouck, J.B., Chambon, P.: {R}ad{G}raph-{XL}: A large-scale expert-annotated dataset for entity and relation extraction from radiology reports. In: Ku, L.W., Martins, A., Srikumar, V. (eds.) Findings of the Association for Computational Linguistics: ACL 2024. pp. 12902--12915. Association for Computational Linguistics, Bangkok, Thailand (Aug 2024). \doi{10.18653/v1/2024.findings-acl.765}

\bibitem{Fischer2023}
Fischer, M., Neher, P., Schüffler, P., Xiao, S., Almeida, Others: Enhanced Diagnostic Fidelity in Pathology Whole Slide Image Compression via Deep Learning, p. 427–436. Springer Nature Switzerland (Oct 2023). \doi{10.1007/978-3-031-45676-3_43}

\bibitem{10.1007/978-3-031-72643-9_13}
Huang, K., Zou, H., Xi, Y., Wang, B., Xie, Z., Yu, L.: Ivtp: Instruction-guided visual token pruning for large vision-language models. In: Leonardis, A., Ricci, E., Roth, S., Russakovsky, O., Sattler, T., Varol, G. (eds.) Computer Vision -- ECCV 2024. pp. 214--230. Springer Nature Switzerland, Cham (2025)

\bibitem{Johnson_Pollard_Berkowitz_Greenbaum_Lungren_Deng_Mark_Horng_2019}
Johnson, A.E.W., Pollard, T.J., Berkowitz, S.J., Greenbaum, N.R., Lungren, M.P., Deng, C.y., Mark, R.G., Horng, S.: Mimic-cxr, a de-identified publicly available database of chest radiographs with free-text reports (Dec 2019), \url{https://www.nature.com/articles/s41597-019-0322-0#citeas}

\bibitem{kaur2022chexprune}
Kaur, N., Mittal, A.: Chexprune: sparse chest x-ray report generation model using multi-attention and one-shot global pruning. Journal of Ambient Intelligence and Humanized Computing  \textbf{14}(6),  7485--7497 (Nov 2023). \doi{10.1007/s12652-022-04454-z}, \url{https://link.springer.com/article/10.1007/s12652-022-04454-z}

\bibitem{proof_of_work}
Kulkarni, P., Kanhere, A.U., Siegel, E., Yi, P.H., Parekh, V.S.: One copy is all you need: Resource-efficient streaming of medical imaging data at scale. CoRR  \textbf{abs/2307.00438} (2023), \url{https://doi.org/10.48550/arXiv.2307.00438}

\bibitem{Li_Nan_Qi_Cai_Zhao_Li_Liu_Wang_Wu_Miao_etal._2024}
Li, J., Nan, Z., Qi, G., Cai, J., Zhao, X., Li, X., Liu, S., Wang, Y., Wu, Y., Miao, X., et~al.: Assessing severity of pediatric pneumonia using multimodal transformers with multi-task learning (Dec 2024), \url{https://pmc.ncbi.nlm.nih.gov/articles/PMC11660274/}

\bibitem{liu2021swintransformerhierarchicalvision}
Liu, Z., Lin, Y., Cao, Y.: Swin transformer: Hierarchical vision transformer using shifted windows (2021), \url{https://arxiv.org/abs/2103.14030}

\bibitem{lu-etal-2022-clinicalt5}
Lu, Q., Dou, D., Nguyen, T.: {C}linical{T}5: A generative language model for clinical text. In: Findings of the Association for Computational Linguistics: EMNLP 2022. pp. 5436--5443. Association for Computational Linguistics, Abu Dhabi, United Arab Emirates (Dec 2022). \doi{10.18653/v1/2022.findings-emnlp.398}

\bibitem{magudia2021trials}
Magudia, K., Bridge, C.P., Andriole, K.P., Rosenthal, M.H.: The trials and tribulations of assembling large medical imaging datasets for machine learning applications. Journal of Digital Imaging  \textbf{34}(6),  1424--1429 (2021). \doi{10.1007/s10278-021-00505-7}

\bibitem{NICOLSON2023102633}
Nicolson, A., Dowling, J., Koopman, B.: Improving chest x-ray report generation by leveraging warm starting. Artificial Intelligence in Medicine  \textbf{144},  102633 (2023). \doi{https://doi.org/10.1016/j.artmed.2023.102633}

\bibitem{REN2024119854}
Ren, H., Jing, F., Chen, Z., He, S., Zhou, J., Liu, L., Jing, R., Lian, W., Tian, J., Zhang, Q., Xu, Z., Cheng, W.: Chexmed: A multimodal learning algorithm for pneumonia detection in the elderly. Information Sciences  \textbf{654},  119854 (2024). \doi{https://doi.org/10.1016/j.ins.2023.119854}

\bibitem{sun2025lvpruningeffectivesimplelanguageguided}
Sun, Y., Xin, Y., Li, H., Sun, J., Lin, C., Batista-Navarro, R.: Lvpruning: An effective yet simple language-guided vision token pruning approach for multi-modal large language models (2025), \url{https://arxiv.org/abs/2501.13652}

\bibitem{Tanida_2023}
Tanida, T., Müller, P., Kaissis, G., Rueckert, D.: Interactive and explainable region-guided radiology report generation. In: 2023 IEEE/CVF Conference on Computer Vision and Pattern Recognition (CVPR). p. 7433–7442. IEEE (Jun 2023). \doi{10.1109/cvpr52729.2023.00718}

\bibitem{ye2024atpllavaadaptivetokenpruning}
Ye, X., Gan, Y., Ge, Y., Zhang, X.P., Tang, Y.: Atp-llava: Adaptive token pruning for large vision language models (2024), \url{https://arxiv.org/abs/2412.00447}

\end{thebibliography}

\end{document}